\documentclass[conference]{IEEEtran}
\IEEEoverridecommandlockouts
\usepackage{cite}
\usepackage{amsmath,amssymb,amsfonts}
\usepackage{algorithmic}
\usepackage{graphicx}
\usepackage{textcomp}
\usepackage{xcolor}
\usepackage{multirow}
\usepackage{authblk}
\def\BibTeX{{\rm B\kern-.05em{\sc i\kern-.025em b}\kern-.08em
    T\kern-.1667em\lower.7ex\hbox{E}\kern-.125emX}}
\begin{document}

\title{HAUR: Human Annotation Understanding and Recognition Through Text-Heavy Images }

\author{Yuchen Yang, Haoran Yan, Yanhao Chen, Qingqiang Wu*, Qingqi Hong*}
\affil{Xiamen University}

\maketitle

\begin{abstract}
Vision Question Answering (VQA) tasks use images to convey critical information to answer text-based questions, which is one of the most common forms of question answering in real-world scenarios. Numerous vision-text models exist today and have performed well on certain VQA tasks. However, these models exhibit significant limitations in understanding human annotations on text-heavy images. To address this, we propose the Human Annotation Understanding and Recognition (HAUR) task. As part of this effort, we introduce the Human Annotation Understanding and Recognition-5 (HAUR-5) dataset, which encompasses five common types of human annotations. Additionally, we developed and trained our model, OCR-Mix. Through comprehensive cross-model comparisons, our results demonstrate that OCR-Mix outperforms other models in this task. Our dataset and model will be released soon .
\end{abstract}


\section{Introduction}
\label{sec:intro}
Question Answering (QA) has always been one of the primary means of human communication. With the advent and development of large language models (LLM), QA tasks have shown immense potential in solving real-world problems. QA tasks in daily life involve various input types, leading to the emergence of Vision Question Answering (VQA) as a method of integrating vision and language. VQA tasks provide a unified question-answering framework for the models, but also pose greater challenges. VQA models aim to address these challenges by answering text-based questions based on images.

\begin{figure}[t]
\centerline{\includegraphics[width=0.5\textwidth]{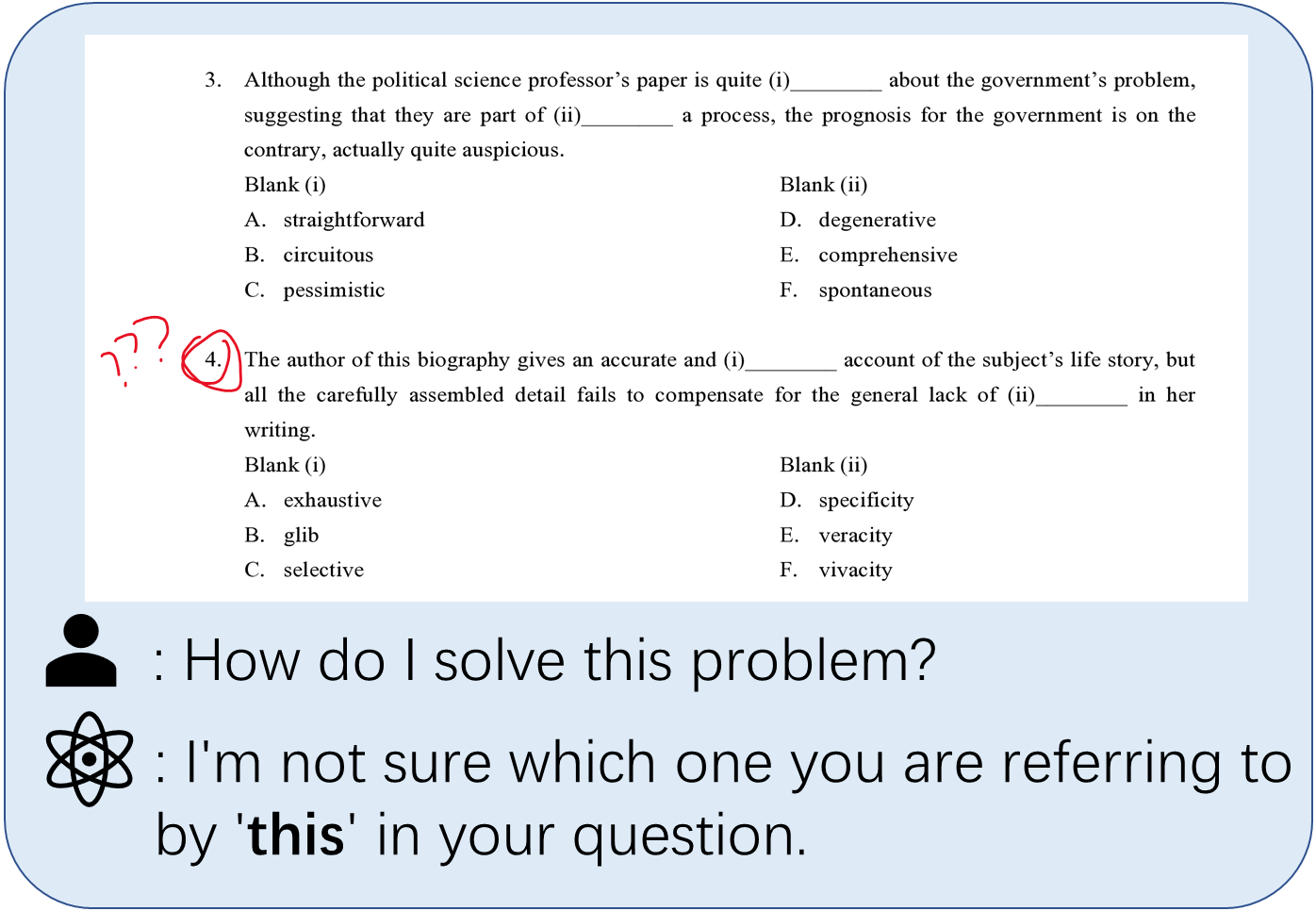}}
\caption{A real-life scenario to explain the motivation behind our task.}
\label{fig:vqa}
\end{figure}

However, existing models still exhibit certain shortcomings in the implementation of VQA tasks. As shown in Figure~\ref{fig:vqa}, consider a real-world scenario: When a student encounters a problem they cannot solve in an exam paper, they might ask the model "How do I solve this question?" At the same time, the student may use simple annotations to indicate the "this" in their query, expecting the model to understand and provide an answer, much like a human teacher would. However, current models struggle to accurately interpret what "this question" refers to in the image. Although the models can solve the problem itself, they fail to pinpoint the specific question indicated by the student, resulting in responses like \textbf{"I don’t know"} or \textbf{answers all the questions}.

To help future models accurately answer questions in the scenarios above, and given the significant progress of current models in reasoning, we propose the \textbf{Human Annotation Understanding and Recognition (HAUR)} task. Our task is defined as taking an input image, understanding and recognizing the content annotated by humans within the image, and outputting it in text form. This task helps multimodal models correctly interpret information indicated by human annotations, enhancing their ability to handle such scenarios effectively. Additionally, it can serve as a module to assist pure language models by converting textual information in images into plain text, enabling them to answer such questions.

To better accomplish the proposed task, we introduce the \textbf{Human Annotation Understanding and Recognition-5 (HAUR-5)} dataset, where the "5" represents the five common types of human annotation styles present we chosen in the text-heavy images within the dataset.

For the proposed task, we introduce the OCR-Mix model. We use Pix2Struct as our base model, as it differs from conventional ViT models by scaling images while preserving their original aspect ratios, rather than resizing them to a square. This approach effectively maintains the natural shape of the text in images, resulting in superior performance on text-heavy images. Additionally, Pix2Struct is pre-trained on a large corpus of documents and web pages, making it well-suited for tasks involving text-heavy images. To further enhance the model's understanding, we incorporate an OCR module to extract all textual content from the images. This extracted text is then used as the model's text input, enabling it to better comprehend the image and generate the required referential information.

In summary, our contributions are as follows:
\begin{figure*}[t]
\centerline{\includegraphics[width=1.0\textwidth]{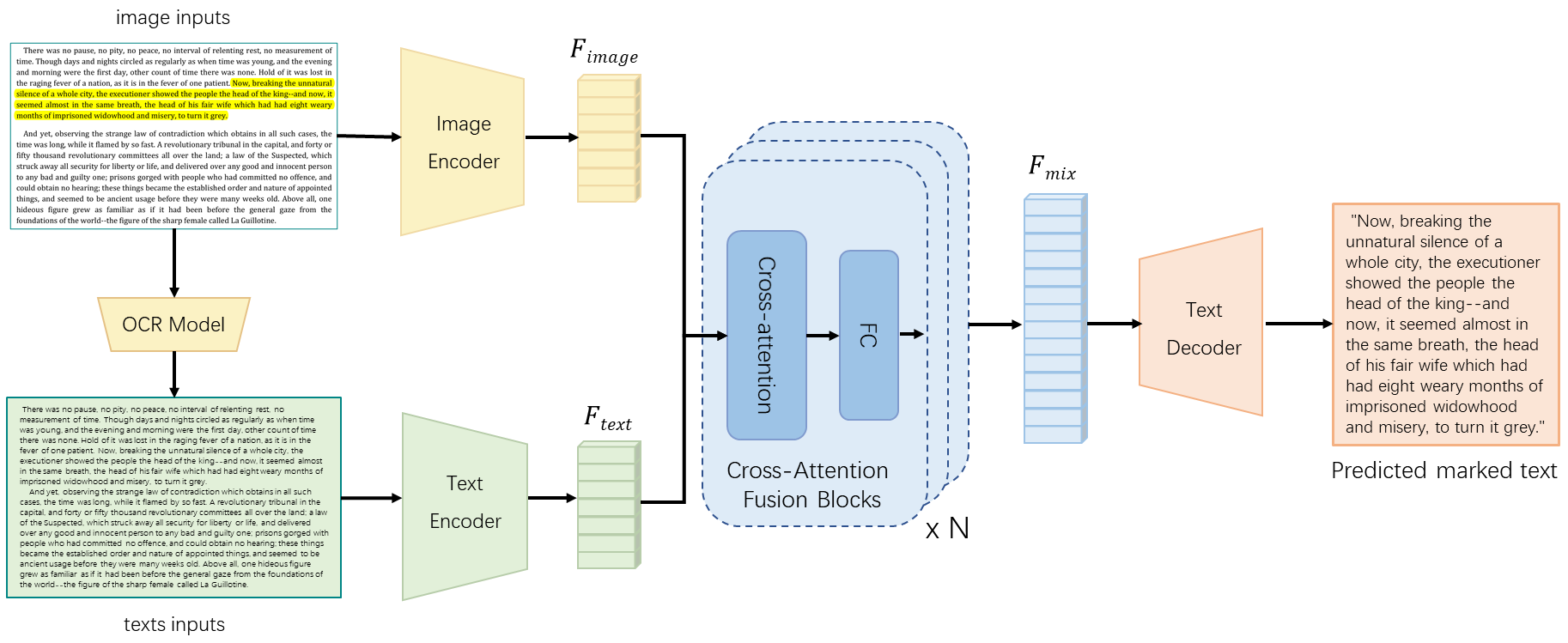}}
\caption{The architecture of our OCR-Mix Model.}
\label{fig:model}
\end{figure*}
\begin{itemize}
    \item We identified an unresolved issue in human annotation recognition in real-world scenarios and propose the HAUR task. This task aims to accurately interpret human annotations and can also assist large language models, enabling them to answer related questions.
    
    \item We propose the HAUR-5 dataset, comprising 37,702 images with five common types of human annotations. We believe that this dataset represents the majority of human annotation scenarios in the real world.
    
    \item We introduce the OCR-Mix model, providing a novel research approach that integrates OCR-based textual features with original image features to generate final outputs, offering valuable insights for future studies.

    \item We conduct extensive experiments, evaluating the performance of various models on our dataset, ranging from specialized VQA models to general-purpose vision models. Our model demonstrates outstanding performance on this task, surpassing existing methods.
\end{itemize}

\section{Related Works}

\subsection{Text-heavy Image Dataset}
Recent research in computer vision has primarily focused on recognition tasks for objects, with relatively few datasets addressing images containing extensive textual information. 
To fill this gap, the Meta team introduced the TextCaps~\cite{sidorov2020textcaps} dataset, which incorporates natural scene text into image recognition.
In Visual Question Answering (VQA), several datasets targeting text-containing images have emerged. The TextVQA~\cite{singh2019towards} and STVQA~\cite{biten2019scene} datasets focus on asking questions about text in natural scenes. RecipeQA~\cite{yagcioglu2018recipeqa} is based on content extracted from recipes. VisualMRC~\cite{tanaka2021visualmrc} collects screenshots of web pages to formulate questions. The TextbookVQA~\cite{li2018textbook} dataset targets pre-printed text and illustrations in textbooks. OCR-VQA~\cite{pham2024viocrvqa} gathers a large collection of book covers and formulates questions based on the text on these covers. DocVQA~\cite{mathew2021docvqa} focuses on extracting and querying different types of information within documents, including body text, titles, tables, and forms. Additionally, InfographicVQA~\cite{mathew2022infographicvqa} targets infographics.

Despite this diversity, these datasets overlook scenarios where preprinted text is annotated with human annotations, which are common in everyday contexts. While existing models can recognize textual content, they often fail to interpret the intent behind human annotations, limiting effective human-machine interaction.

\subsection{VQA Models for Text-heavy Images}
Text detection and recognition in natural images form the foundation of our research. With the success of Transformer architectures in Natural Language Processing (NLP), studies have explored integrating image and text features using various Transformer configurations, such as Meta's M4C~\cite{hu2020iterative} and Yang et al.'s TAP~\cite{yang2021tap}.
Later models use simple text and image encoders combined with more complex interaction modules for efficient training, as discussed in ViLT~\cite{kim2021vilt}. For example, PaLi~\cite{chen2022pali} and PaLi-3~\cite{chen2023pali} employ large Transformer Encoder-Decoder structures, while ScreenAI~\cite{baechler2024screenai}, based on Pix2struct~\cite{lee2023pix2struct}, adapts image preprocessing to preserve aspect ratio and tile consistency, which informs our approach.
The rise of large language models (LLMs) has led to the development of multimodal LLMs that unify image, audio, and text tasks, such as LLava~\cite{liu2024llavanext,liu2023improvedllava}, LLama-Vision, and Qwen-2-VL~\cite{Qwen-VL,Qwen2VL}. 

In our experiment, we will also evaluate the performance of different multimodal models on our task.

\section{Methods}
As illustrated in Figure~\ref{fig:model}, the overall process of the model in this study consists of the following stages: First, advanced OCR technology is used to extract textual information from the input images, which is then fed into a language model to generate text features. Second, a pretrained visual encoder is employed to extract image features. Finally, a multi-layer cross-attention mechanism integrates the multimodal features of text and images, and a decoder generates the final prediction.

\subsection{Data Preprocessing}
We use a robust OCR model to extract text from each image and save the recognized text as a .txt file, which serves as the textual input, denoted as $T$, for the model. Notably, this step does not depend on any specific OCR model, ensuring the approach's generalizability.

\subsection{Model Architecture}
\subsubsection{Text Feature Encoding}
In the text feature extraction component, we use a text encoder to embed the extracted text \( T \), transforming it into high-dimensional feature vectors \( F_{{text}} \):
\begin{equation}
F_{{t}} =\mathbb{E}_{text}(T).
\end{equation}

\subsubsection{Image Feature Encoding}
For image feature, we used the pre-trained image encoder of Pix2Struct to convert the image \( I \) into a feature vector \( F_{{image}} \) and we set the maximum number of patches to the default value of 2048: 
\begin{equation}
F_{{i}} = \mathbb{E}_{image}(I).
\end{equation}

\subsubsection{Multimodal Feature Fusion}
The image features \( F_{\text{image}} \) and text features \( F_{\text{text}} \) are fed into a fusion network composed of multiple layers of Cross-Attention Modules (CAM). The cross-attention mechanism is computed using the following formula:
\begin{IEEEeqnarray}{l}
F_\text{mixed}^{n}  = 
\text{softmax} \left(
    \frac{
        Q^{n}(F_{ca}^{n}) K^{n}(F_{i})^\top
    }{
        \sqrt{d_k}
    }
\right)  V^{n}(F_{i}) 
\end{IEEEeqnarray}

\begin{IEEEeqnarray}{rCl}
F_{ca}^0 & = & F_{t}
\end{IEEEeqnarray}

Where \( Q \), \( K \), and \( V \) represent Query, Key, and Value projection layers , respectively, and  \( d_k \) is the dimension of the Key vector.

After the cross-attention mechanism, a Fully Connected (FC) layer is applied to transform the fused features. The transformation is mathematically expressed as:
\begin{equation}
F_{ca}^{n+1} = W^{n+1}\times F^{n}_{mixed} + b^{n+1}
\end{equation}

\subsubsection{Decoding and Generation}
Finally, the fused features \( F^{N}_{ca} \) are fed into the decoder to generate the target text. The objective is to produce the logits of the output text from the fused features:
\begin{equation}
Logits = \mathbb{D}_{text}(F^{N}_{ca}).
\end{equation}

\section{Experiment}
\subsection{Dataset}
\begin{figure}[t]
\centerline{\includegraphics[width=0.5\textwidth]{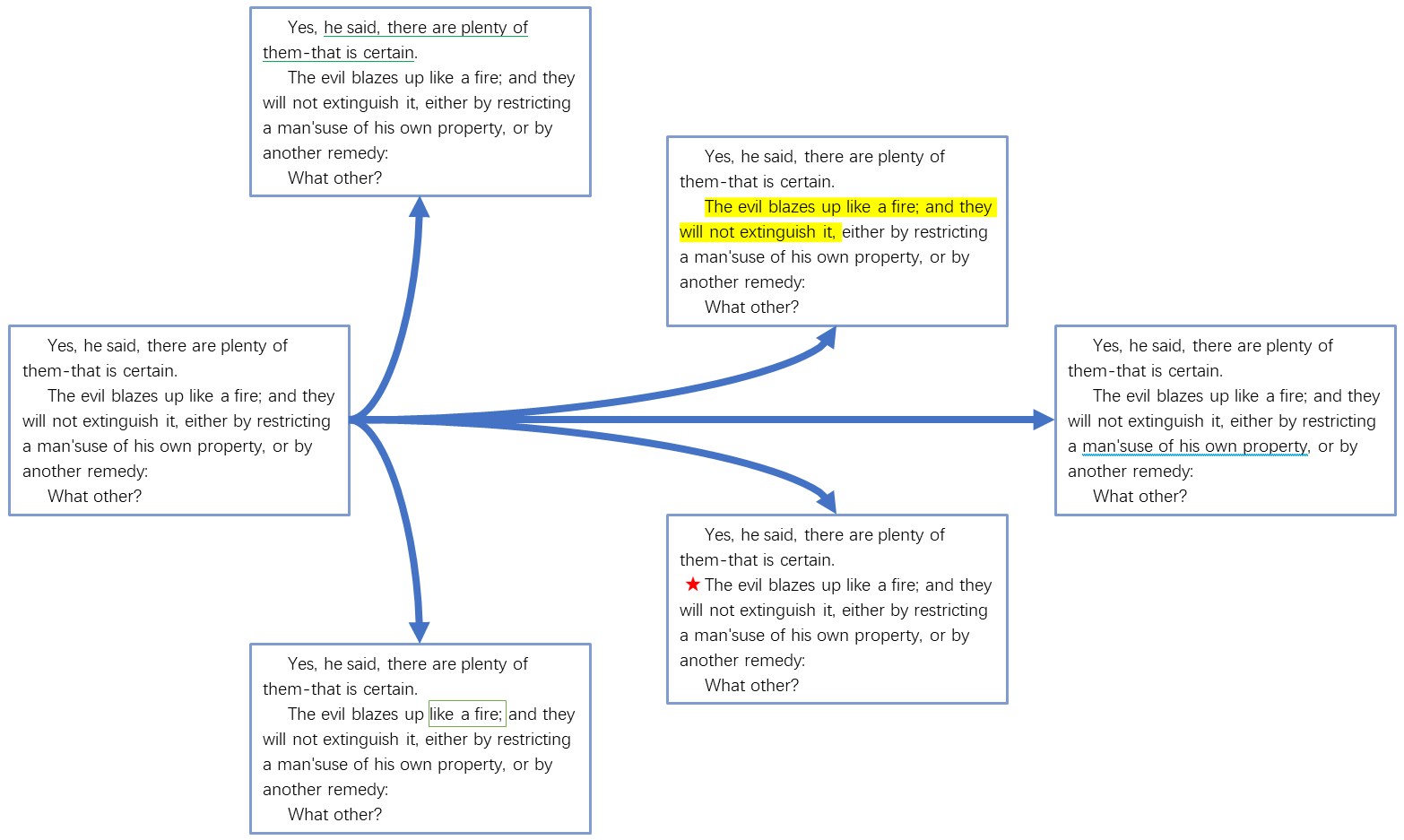}}
\caption{A simple image showcasing the contents of our HAUR-5 dataset: The dataset includes five common types of human annotations.}
\label{fig:dataset}
\end{figure}

We selected 16 classic novels from the Project Gutenberg website that are no longer under copyright protection. The content of these novels was divided into text chunks of varying sizes, and five types of annotations were applied to these chunks: \textbf{Highlight}, \textbf{Underline}, \textbf{Squiggly} underline, \textbf{Rectangular} bounding box, and special symbols at the beginning of paragraphs (\textbf{Paragraph Marking}), as shown in the simple example in Figure~\ref{fig:dataset}.

For highlight, underline and squiggly underline annotations, a random sentence was selected from the text chunks, which were then converted into PDFs for annotation. To simulate human imprecision, noise was introduced by adding extra characters from the beginning or end of adjacent sentences. After completing the annotations, the PDFs were converted into images. 

For the rectangular box annotations, text chunks were directly converted into images. OCR technology was used to identify the coordinates of randomly selected sentences, and rectangular boxes were drawn around the corresponding regions to complete the annotations.

For paragraph marking annotations, a random paragraph was selected, and a special star symbol was inserted at the beginning of the paragraph. This symbol occupies two whitespace characters and does not cause any ambiguity with other words. OCR was then employed to locate the symbol, which was subsequently covered with a white block, and the annotation was drawn.

Finally, the annotated images were cropped to remove unnecessary blank areas 
% using a tool program fine-tuned with DocLayNet~\cite{doclaynet2022} and YOLOv10~\cite{wang2024yolov10}
to form the dataset. The dataset includes 7,750 images with underline annotations, 7,750 with squiggly annotations, 7,749 with rectangular box annotations, 7,750 with highlight annotations, and 6,703 with paragraph marking annotations. The lower number of special symbol samples is due to the exclusion of images containing only a single paragraph.
Due to space constraints, additional analyses related to the dataset are provided in the appendix.

\subsection{Experiment Setup}
\subsubsection{\textbf{Metrics}}
Like most VQA methods, we used the Average Normalized Levenshtein Similarity (ANLS) \cite{biten2019scene} and Accuracy (ACC) metrics to evaluate the performance of various models on our task. 
Among these two metrics, ACC requires the answer to be exactly the same as the Ground Truth to be considered correct, while ANLS only considers the answer completely incorrect when the distance between the two exceeds a threshold. Otherwise, ANLS provides a smooth response that can gracefully capture errors.
The formula for ANLS is as follows:
\begin{IEEEeqnarray}{rCl}
\text{ANLS} & = & \frac{1}{N} \sum_{i=1}^N \left( \max_j s(a_{ij}, o_q) \right)
\end{IEEEeqnarray}

\begin{IEEEeqnarray}{rCl}
s(a_{ij}, o_q) & = &
\begin{cases}
1 - \text{NL}(a_{ij}, o_q), & \text{if } \text{NL}(a_{ij}, o_q) < \tau, \\
0, & \text{if } \text{NL}(a_{ij}, o_q) \geq \tau.
\end{cases}
\end{IEEEeqnarray}

Where $N$ is the total number of questions in the dataset, $M$ is the total number of GT answers per question, $a_{ij}$ are the ground truth answers where $i = \{0, \dots, N\}$, and $j = \{0, \dots, M\}$, and $o_q$ is the network's answer for the $i^\text{th}$ question $q_i$. ${NL}(a_{ij}, o_q)$ is the normalized Levenshtein distance between the strings $ a_{ij}$ and $o_q$. In our experiments, we set the threshold  $\tau$ of ANLS to 0.5 following the original setting \cite{biten2019scene}. 

% \subsubsection{Model Configuration}
\begin{table*}[t]
\caption{The Performance of Our Model Compared to Various Other Models on Our Dataset}
\begin{center}
\begin{tabular}{|c|cc|cc|cc|cc|cc|cc|}
\hline
\textbf{Models} & \multicolumn{2}{|c|}{\textbf{Average Score}} & \multicolumn{2}{|c|}{\textbf{Highlight}} & \multicolumn{2}{|c|}{\textbf{Underline}} & \multicolumn{2}{|c|}{\textbf{Squiggly}} & \multicolumn{2}{|c|}{\textbf{PM}} & \multicolumn{2}{|c|}{\textbf{Rect}} \\ 
& \textbf{\textit{ACC}} & \textbf{\textit{ANLS}} & \textbf{\textit{ACC}} & \textbf{\textit{ANLS}} & \textbf{\textit{ACC}} & \textbf{\textit{ANLS}} & \textbf{\textit{ACC}} & \textbf{\textit{ANLS}} & \textbf{\textit{ACC}} & \textbf{\textit{ANLS}} & \textbf{\textit{ACC}} & \textbf{\textit{ANLS}} \\ 
\hline
LLAVA:7B & 0.00 & 0.28 & 0.00 & 0.71 & 0.00 & 0.19 & 0.00 & 0.18 & 0.00 & 0.10 & 0.00 & 0.19 \\
PailGemma:3B & 0.40 & 0.42 & 1.37 & 0.03 & 0.17 & 0.04 & 0.08 & 0.02 & 0.00 & 0.13 & 0.39 & 0.01 \\
Minicpm-V:8B & 9.37 & 0.05 & 12.77 & 0.06 & 2.19 & 0.02 & 1.16 & 0.01 & 2.54 & 0.08 & 31.14 & 0.01 \\
LLAMA-V:11B & 0.09 & 0.04 & 0.00 & 0.66 & 0.00 & 0.54 & 0.00 & 0.41 & 0.45 & 0.20 & 0.00 & 0.27 \\
Qwen2-VL-7B-Instruct & 0.00 & 0.04 & 0.00 & 0.23 & 0.00 & 0.06 & 0.00 & 0.04 & 0.00 & 0.17 & 0.00 & 0.38 \\
GLM-4 (web) & 32.06 & 0.78 & 54.89 & 0.95 & 24.89 & 0.79 & 15.33 & 0.76 & 38.37 & 0.77 & 25.98 & 0.63 \\
Qwen2.5 (web) & 44.53 & 0.88 & 65.72 & 0.97 & 43.22 & 0.89 & 25.68 & 0.87 & 15.21 & 0.74 & 77.28 & 0.92 \\
GPT-4o (web) & 46.78 & 0.90 & 71.38 & 0.99 & 56.74 & 0.96 & 25.97 & 0.84 & 15.23 & 0.89 & 67.36 & 0.79 \\
\hline
Pix2Struct(0-shot) & 7.76 & 0.17 & 16.01 & 0.07 & 5.34 & 0.03 & 1.43 & 0.04 & 0.51 & 0.05 & 16.72 & 0.02 \\
Pix2Struct(fine-tuned) & 72.41 & 0.95 & 81.12 & 0.96 & 75.57 & 0.96 & 75.10 & 0.96 & 45.12 & 0.87 & 87.12 & 0.99 \\
\hline
Mix-OCR (Ours):1.31B & \textbf{86.47} & \textbf{0.98} & \textbf{91.35} & \textbf{0.99} & \textbf{90.71} & \textbf{0.99} & \textbf{91.22} & \textbf{0.99} & \textbf{61.23} & \textbf{0.96} & \textbf{99.61} & \textbf{0.99} \\ 
\hline
\multicolumn{13}{l}{All web models are commercial models with undisclosed parameter counts, but their parameter sizes are all above 100B.} \\
\multicolumn{13}{l}{PM stands for Paragraph Marking.}
\end{tabular}
\label{tab:res}
\end{center}
\end{table*}

\subsubsection{\textbf{Compared Methods}}
In extensive experiments, we found that since most VQA models do not have OCR capabilities, their experimental results are not competitive and have no reference value. Therefore, in our comparative experiments, we selected Pix2Struct, a VQA models with OCR ability. We show both zero-shot and fine-tuned scenarios. We also select multimodal LLMs, such as LLAVA, Qwen2-VL, Minicpm, PailGemma, and LLAMA 3.2-vision, as well as larger models deployed on the web, such as Qwen2.5, ChatGLM4, and GPT-4o.

\subsubsection{\textbf{Implement Details}}
For our experiments, we used the PaddleOCR model as a reference.
The image encoder in our model utilizes the encoder component of Pix2Struct-base, which has been pre-trained on webpages with abundant textual content, making it well-suited for our task. For the text encoder, we employed a learnable embedding module with a vocabulary size of 50,244 and a hidden state size of 768. The text decoder in our model uses the decoder component of Pix2Struct-base, which, as described in its paper, is essentially a T5 encoder-decoder module.

We divided the dataset into 80\% for training, 10\% for validation, and 10\% for testing. The training process used the AdamW optimizer with an initial learning rate of 1e-5, along with a CosineAnnealingLR scheduler. The optimization objective was the Cross-Entropy Loss function. Training was conducted over 50 epochs, and the best-performing model on the validation set was selected for testing. The fusion module in our model was configured with \( N=4 \). All training was carried out on a single NVIDIA RTX 4090 GPU.

\subsection{Main Result}
We conducted tests on various published open-source models for our task, shown in Table~\ref{tab:res}.
% As shown in Figure~\ref{fig:query}, 
For the VQA model, we directly pose questions, while for the multimodal large language model, we use few-shot prompts. Due to space limitations, query and prompt are shown in the appendix.
% For VQA models, the question we posed is: "What's the text content marked in the image?"

% For multimodal large language models, we used a few-shot prompt approach:
% "What's the text content marked in the image? Recognize the marked text in the image. Respond with only the recognized text or ' ' if you cannot identify any text.
% For example, if an image contains the text: 'when I live in such a world of fools as this? Merry Christmas!' and the marked text is 'Merry Christmas!', if you recognize it, respond with 'Merry Christmas!', otherwise respond with ' '."

% \begin{figure}[t]
% \centerline{\includegraphics[width=0.35\textwidth]{query.png}}
% \caption{Query in the VQA Model and Prompt in the VLLM}
% \label{fig:query}
% \end{figure}

The experimental results show that our model demonstrates outstanding performance across all tasks, surpassing all other models. For every task, the ANLS metric of our model exceeds 0.95, and in four of five annotation methods, the ACC exceeds 90\%, showcasing exceptional capability.
Compared to GPT-4o, the closest competitor, our model achieves an ACC improvement ranging from 1.28 times to 4 times and an ANLS improvement of 8\% to 25\% across different tasks, further solidifying its leading position in HAUR tasks. 
Further analysis reveals that models perform best on rectangular box annotations, which involve shorter text, and struggle more with paragraph markings, which are longer and more complex. Despite lower accuracy, the ANLS remains high for paragraph markings. Annotations like highlights, underlines, and rectangles are visually distinct and easier for models to recognize, while paragraph markings and squiggly underlines are more complex, requiring advanced feature extraction. Compared to the fine-tuned Pix2Struct model, our model offers greater advantages by incorporating text input and fusing the features of both modalities through the fusion module. This approach allows for better recognition of annotated text in the image.

Interestingly, models like LLAVA, LLAMA3.2-V, and Qwen2-VL perform worse on our dataset than some smaller models. We believe this is because they likely didn't encounter text-heavy scenarios or specific annotation styles during pre-training, or their architectures lack the necessary text recognition capabilities. In contrast, our base model, Pix2Struct, is specialized in structured document parsing. Despite having fewer parameters, this specialization allows it to handle annotation recognition tasks more effectively, even outperforming larger multimodal models in some cases.

Super-large commercial models like GPT-4o-web and Qwen2.5-web show a significant performance gap compared to other models, mainly due to their massive parameter scales. Their zero-shot capabilities greatly surpass smaller models, enhancing their comprehension abilities. Their pre-training likely involved annotation tasks, allowing them to perform well on certain annotations, such as highlights and rectangles. Although their ACC is low, their ANLS remains relatively high. We found this discrepancy is due to their imprecision in judgment and difficulty handling noise from handwritten annotations, which leads to redundant context being output. This results in incorrect classifications for ACC, but since the redundant context doesn’t push the prediction past the ANLS threshold, it doesn’t count as entirely incorrect, leading to a higher score.

\subsection{Ablation Experiments}
\subsubsection{\textbf{Different Text Encoder}}
We experimented with different text encoder modules, including the T5 encoder, BERT encoder, and a learnable embedding module. The experimental results are shown in Table~\ref{tab:encoder}. Interestingly, we found that the pre-trained T5 and BERT encoders underperformed compared to the simpler learnable embedding module. This can be attributed to the relatively small text range and limited vocabulary in our task, results in a sparse attention to the text. Large pre-trained models may introduce unnecessary noise, ultimately hindering overall performance. In contrast, the learnable embedding module is specifically designed for this task, with its training process entirely reliant on task-specific data. As a result, it demonstrates significantly better adaptability and performance.

\subsubsection{\textbf{Different Fusion Module Layers}}
\begin{figure}[t]
\centerline{\includegraphics[width=0.5\textwidth]{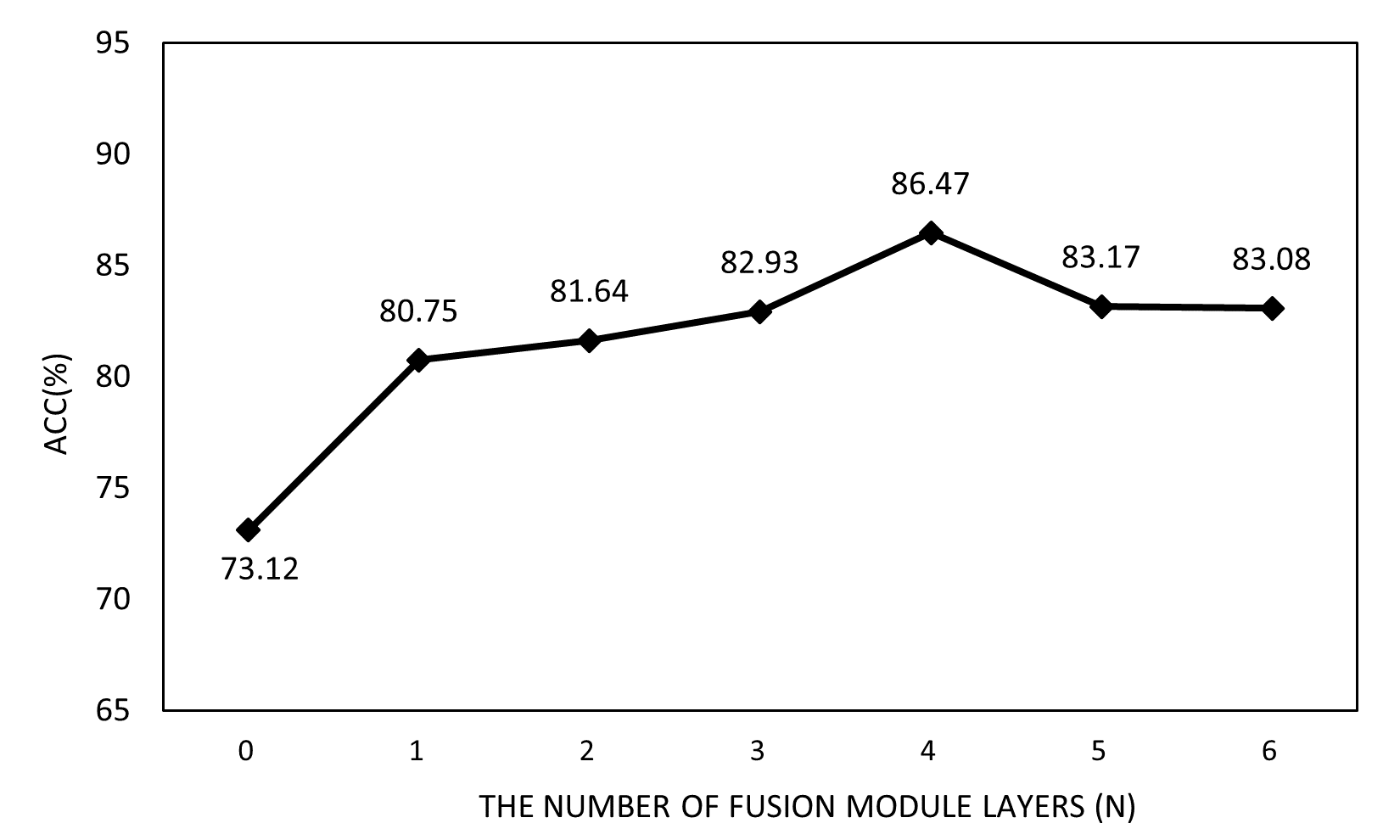}}
\caption{Acc with Different Fusion Module Layers}
\label{fig:layers}
\end{figure}

We experimented with the impact of different numbers of fusion module layers ($N$) on the performance of our model, with results shown in Figure~\ref{fig:layers}. For $N=0$, we concatenated the text and image features and input them into the decoder via a learnable FC layer. The results indicate that the model performs best when $N=4$. As $N$ increases further, performance declines.
We believe that for small values of $N$, increasing the number of layers helps the fusion module better combine text and image features, improving performance. However, when $N$ exceeds a certain threshold, the query, set to the output of the previous layer, becomes increasingly dominated by fixed image features. This reduces the proportion of text features in the fused features, while excess redundant information from the image features adds interference to the decoder, ultimately degrading performance. Therefore, we chose $N=4$ as the optimal number of fusion module layers for OCR-Mix.
\begin{table}[t]
\caption{Acc with Different Text Encoders}
\begin{center}
\begin{tabular}{|c|c|}
\hline
\textbf{Text Encoder} & \textbf{ACC (\%)} \\
\hline
BERT-encoder & 81.96 \\
T5-encoder & 83.64 \\
Learnable Embedding (Ours) & 86.47 \\
\hline
% \multicolumn{2}{l}{This table shows the performance of different text encoders.} % 可选脚注
\end{tabular}
\label{tab:encoder}
\end{center}
\end{table}

% \begin{table}[t] 
% \caption{Acc with Different Fusion Layers}
% \begin{center}
% \begin{tabular}{|c|c|c|}
% \hline
% \textbf{N} & \textbf{ACC (\%)} \\ 
% \hline
% 0     & 67.96  \\ 
% 1     & 83.98  \\
% 2     & 87.05  \\
% 3     & 86.82  \\
% 4     & 83.26  \\
% 5     &        \\
% 6     &        \\
% \hline
% % \multicolumn{3}{l}{$^{\mathrm{a}}$This table shows the performance of different fusion layers in terms of GFLOPS and accuracy.} % 添加脚注
% \end{tabular}
% \label{tab:layers}
% \end{center}
% \end{table}

\begin{table}[t]
\caption{The Effects of Different Cross-Attention Configurations on Model Performance}
\begin{center}
\begin{tabular}{|c|c|}
\hline
\textbf{Configuration} & \textbf{ACC (\%)} \\ 
\hline
Image-Text-Text & 67.96 \\ 
Text-Image-Image & 86.47 \\
\hline
\end{tabular}
\label{tab:qkv}
\end{center}
\end{table}

\subsubsection{\textbf{Different Cross-Attention Configuration}}
We experimented with different sources for the query (Q), key (K), and value (V) in the fusion module with cross-attention, as shown in Table~\ref{tab:qkv}. Specifically, 'Image-Text-Text' means the Q comes from image features, while K and V come from text features. Conversely, 'Text-Image-Image' means Q is from text features, and K and V are from image features.
The results show that using text features as the query source and image features as the keys and values yields better performance than the reverse configuration. This is because the task involves selecting relevant parts of the text based on image features and generating the output accordingly. The configuration with text features (and mixed features) as the query source, and image features as the keys and values, better aligns with the task. In the multi-layer cross-attention module, the model can enhance the weight of relevant text features based on image features while suppressing irrelevant text, effectively extracting the required information.
\subsection{In-wild Data Result}

\begin{figure}[t]
\centerline{\includegraphics[width=0.5\textwidth]{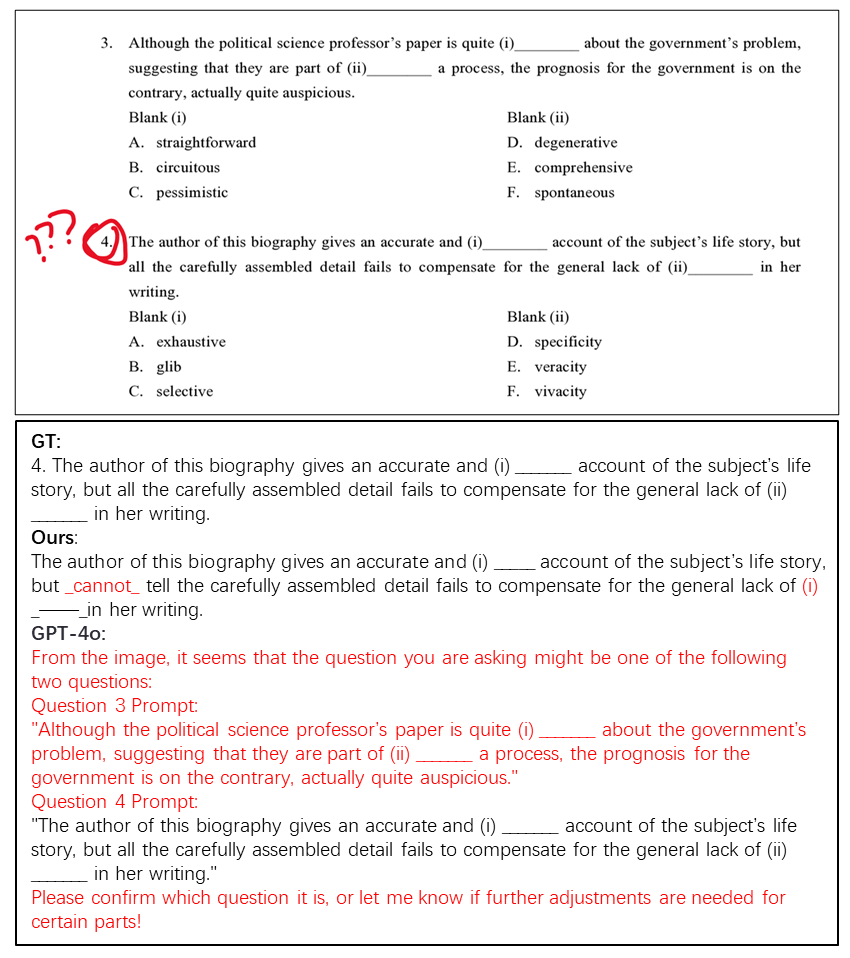}}
\caption{Experiment with human-labeled images in real-world scenarios: Red text represents the difference between the model’s prediction and the actual value.}
\label{fig:in-wild}
\end{figure}
We also tested the performance of our model on annotations in real-world scenarios, with one example shown in Figure~\ref{fig:in-wild}. Additional examples are provided in the appendix. The results indicate that, while GPT can accurately recognize text within images, it struggles to identify and understand human annotations. Moreover, when human-drawn annotations contain slight errors, GPT has difficulty recognizing them accurately. In contrast, our model, with minimal errors, successfully understands the human annotations and generates accurate prompts. 
This demonstrates that existing models still face limitations in addressing the demands of real-world contexts and practical applications. In contrast, our model not only excels on the designed dataset but also shows strong generalization capabilities in real-world scenarios.

\section{Conclusion and Future Works}
We propose a novel task, Human Annotation Understanding and Recognition (HAUR), aimed at enabling models to comprehend and recognize human annotations, especially in text-heavy images. To support this task, we introduce the HAUR-5 dataset, a foundational resource for advancing models in this domain.
Additionally, we present the OCR-Mix model, which integrates image and text features extracted from images, achieving exceptional performance on the HAUR-5 dataset. Extensive experiments show that OCR-Mix outperforms other models on the HAUR task. Moreover, OCR-Mix can serve as an external tool for vision-language large models, enhancing their ability to handle real-world VQA tasks.
We hope our work inspires similar efforts and stimulates further research into understanding images with complex human annotations and embedded text.

Although our work can be generalized to real wild data through artificially simulated datasets, it should be noted that the performance of the model may be different when applied to real-world data due to the differences between simulated conditions and real conditions. Therefore, our next step will be to collect and annotate real-world data and further improve the robustness and generalizability of the model.

\bibliographystyle{IEEEbib}
\bibliography{icme2025references}
\clearpage
\section*{Appendix}
% 附录部分
\subsection{Dataset Collection}
\begin{figure}[htbp]
\centerline{\includegraphics[width=0.5\textwidth]{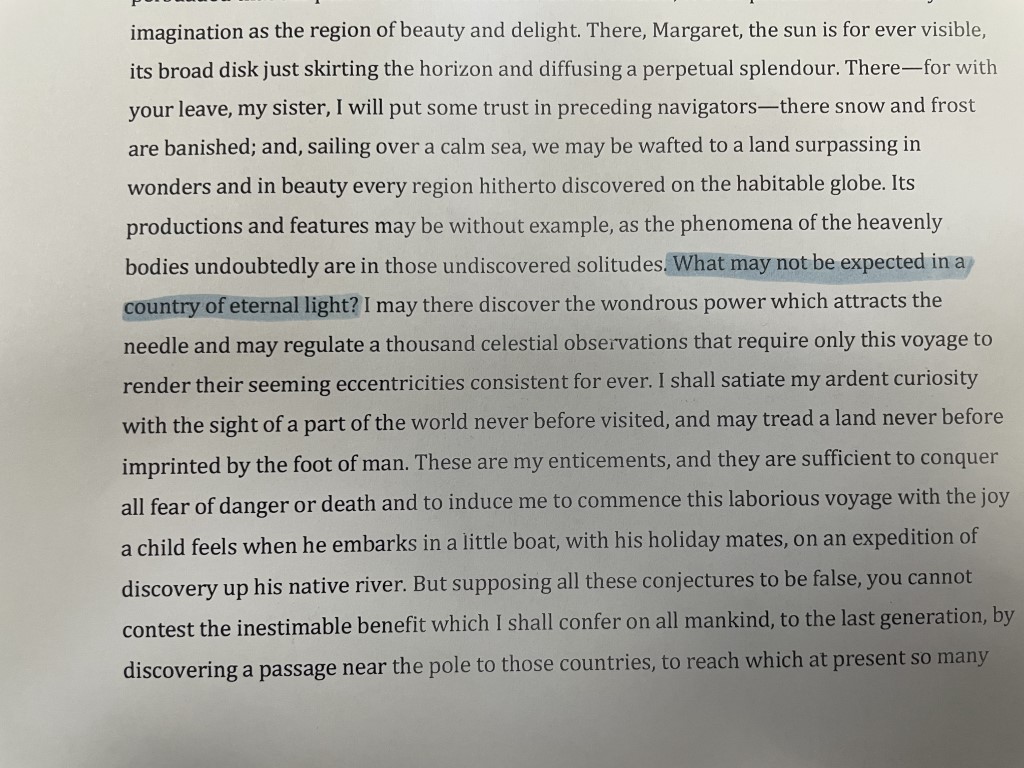}}
\caption{An example of human annotation in real life.}
\label{fig:realLife}
\end{figure}

During the dataset creation process, we downloaded the TXT files of 16 novels from Project Gutenberg, all of which have passed the copyright expiration. We manually organized the format of these texts, including extracting titles, formatting paragraphs, and adjusting line breaks. Subsequently, we used the docx-python library to segment the text into chunks, ensuring that the token count of each chunk did not exceed 500. This limitation was based on our understanding of real-world scenarios: even if the entire text on a page is long, people typically do not photograph the entire page but instead focus on capturing key content and its relevant context, as shown in the Figure~\ref{fig:realLife}.

\begin{figure}[htbp]
\centerline{\includegraphics[width=0.5\textwidth]{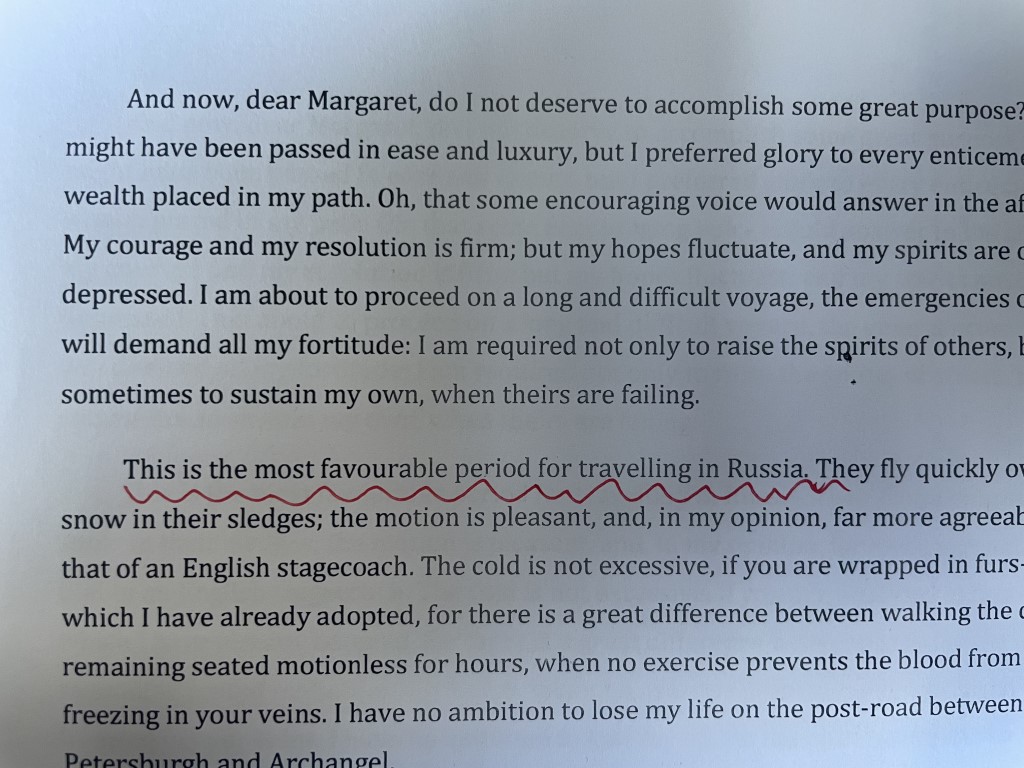}}
\caption{An example of human annotation with errors.}
\label{fig:errors}
\end{figure}
Afterward, we applied five types of annotations to all text chunks, as described in the main text of the paper. Specifically, for the Highlight, Straight underline, and Squiggly underline annotations, we used the PyMuPDF library to mark the text in PDF files. It is important to note that, to simulate the potential errors humans might make while annotating, we introduced noise during the marking process for these three methods by randomly extending the annotations to the end of the previous sentence or the end of the next sentence, as shown in the Figure~\ref{fig:errors}. The PM method is slightly more complex; we first manually selected chunks that contained multiple paragraphs and then applied the annotation. For the Rect annotation, we directly used the OpenCV library to draw rectangular boxes in the images.

Finally, we pass our image through a fine-tuned python program tool of YOLOv10 and DocLayout to crop the extra blank areas and get our image.
% \clearpage
\subsection{Dataset Distribution}

\begin{figure}[htbp]
\centerline{\includegraphics[width=0.5\textwidth]{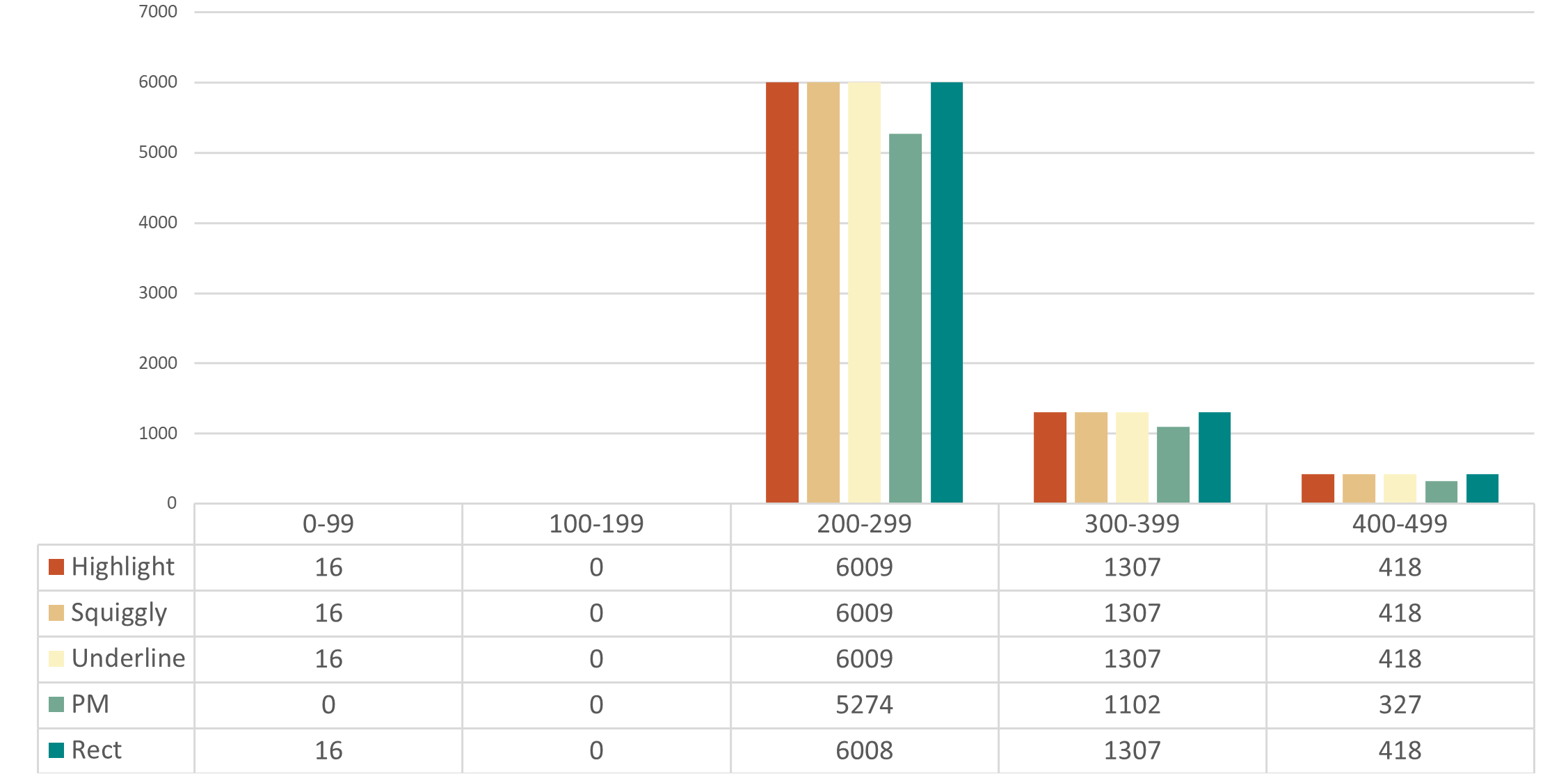}}
\caption{Full Text Length Distribution}
\label{fig:full_texts_length}
\end{figure}

\begin{figure}[t]
\centerline{\includegraphics[width=0.5\textwidth]{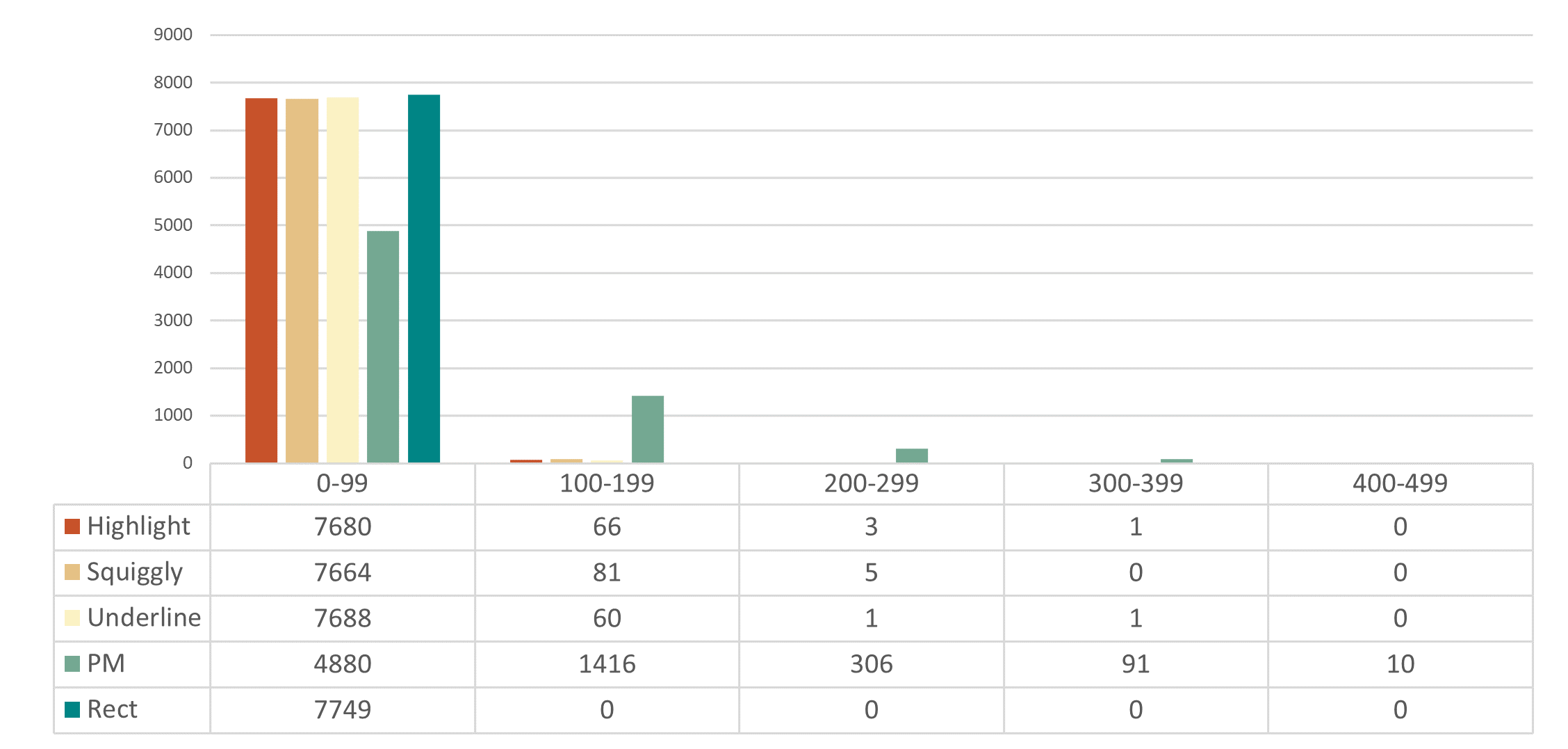}}
\caption{Ground Truth Length Distribution}
\label{fig:ground_truth_length}
\end{figure}

The length of all text in the images of our dataset and the corresponding ground truth lengths are shown in the Figure~\ref{fig:full_texts_length} and Figure~\ref{fig:ground_truth_length}.

Regarding the full text length, during dataset creation, we imposed a restriction that the total number of tokens in the text of each image should not exceed 500. The results indicate that the distribution of text length is concentrated between 200 and 300 tokens. Among the four annotation methods—Highlight, Squiggly, Underline, and Rect—there is a high degree of consistency in text length, as they all derive from the same text chunks. In contrast, the PM method differs significantly in terms of text length distribution due to the exclusion of text chunks with only one paragraph, which results in a distribution that differs notably from the other methods.

As for the ground truth length, all five methods exhibit a large proportion of text lengths in the 0-100 token range. This is because for Highlight, Squiggly, and Underline, typically only one sentence is selected for annotation, while for Rect, the selection usually involves 1-5 tokens. In the case of the PM method, the length of the selected text depends on the length of the paragraph. Given the restriction that no chunk exceeds 500 tokens and the requirement for multiple segments to distinguish annotated and non-annotated paragraphs in a single image, even the PM method does not produce excessively long segments as ground truth.

\subsection{Query and Prompt}
\begin{figure}[htbp]
\centerline{\includegraphics[width=0.35\textwidth]{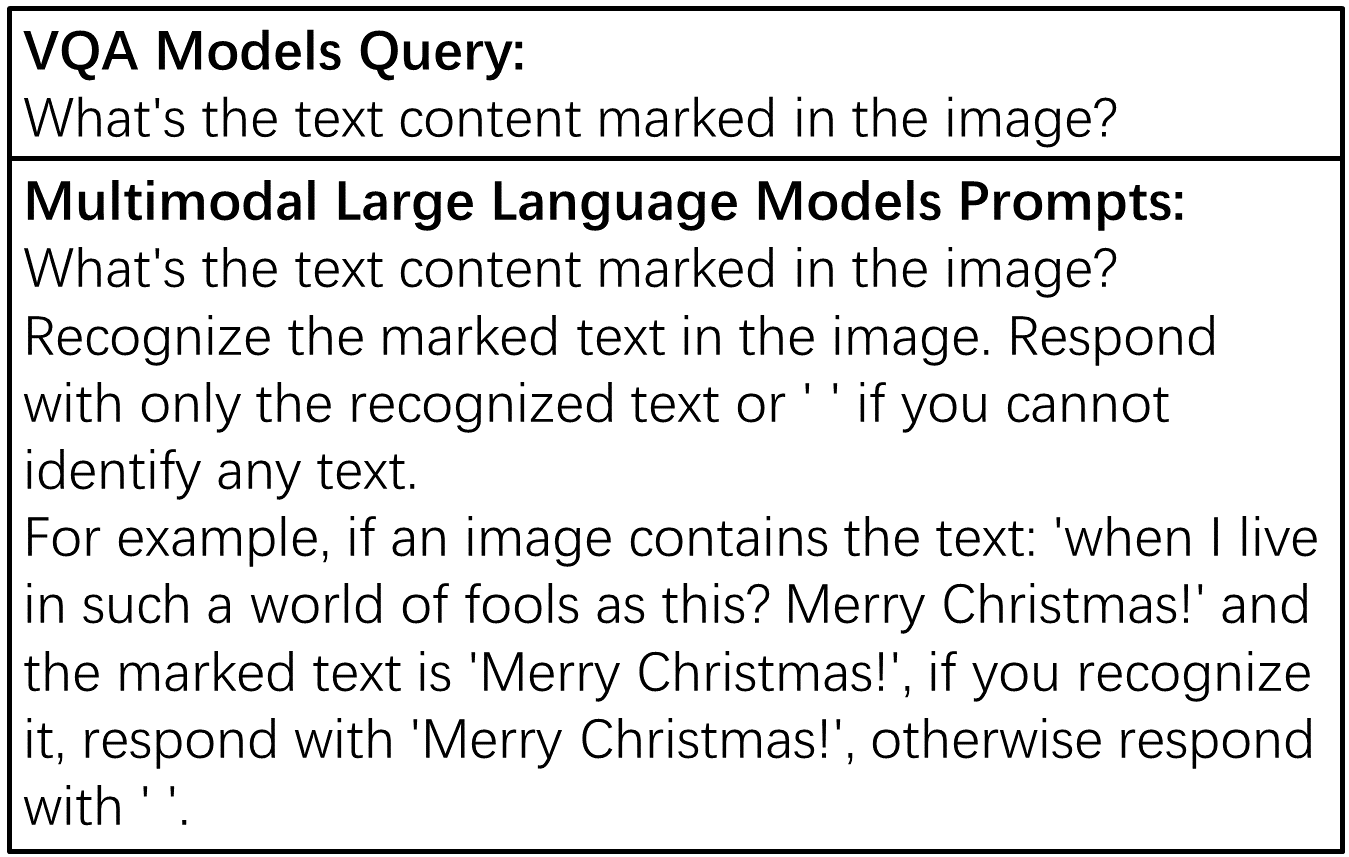}}
\caption{Query in the VQA Model and Prompt in the VLLM}
\label{fig:query}
\end{figure}

As shown in Figure~\ref{fig:query}, for the VQA model, we directly pose questions, while for the multimodal large language model, we use few-shot prompts. Although we used a few-shot prompt, we found that the model occasionally outputs a prefix like 'The mark in this picture is:'.

\subsection{In-wild Data Results}
\begin{figure}[htbp]
\centerline{\includegraphics[width=0.5\textwidth]{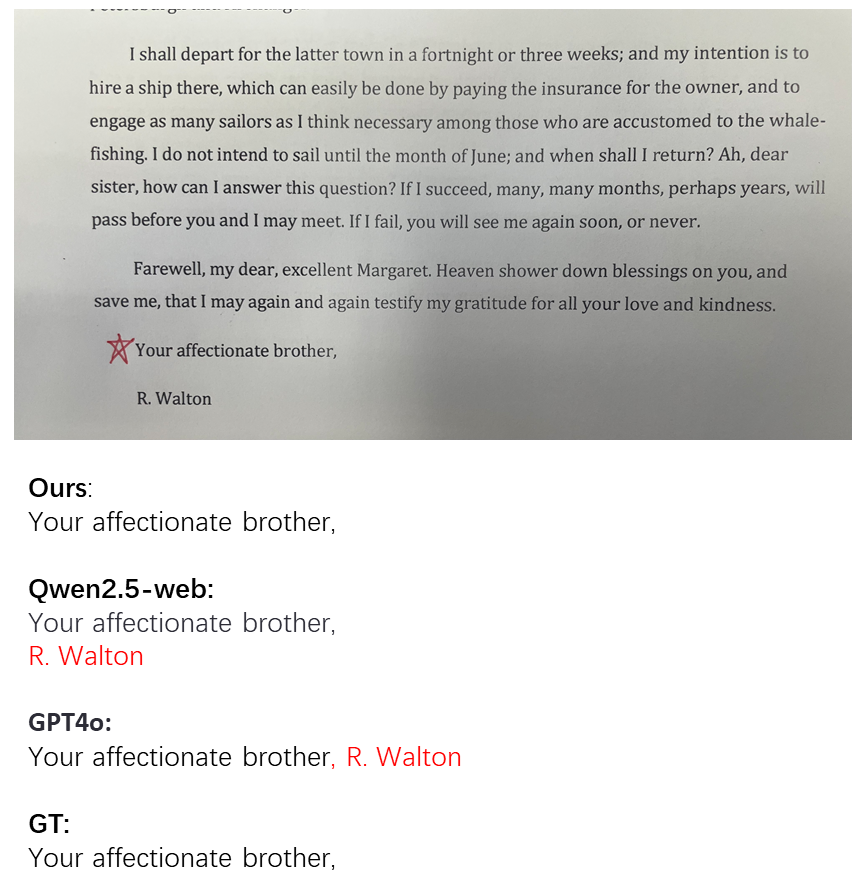}}
\caption{A example of in-wild data results.}
\label{fig:wild}
\end{figure}

\begin{figure}[htbp]
\centerline{\includegraphics[width=0.5\textwidth]{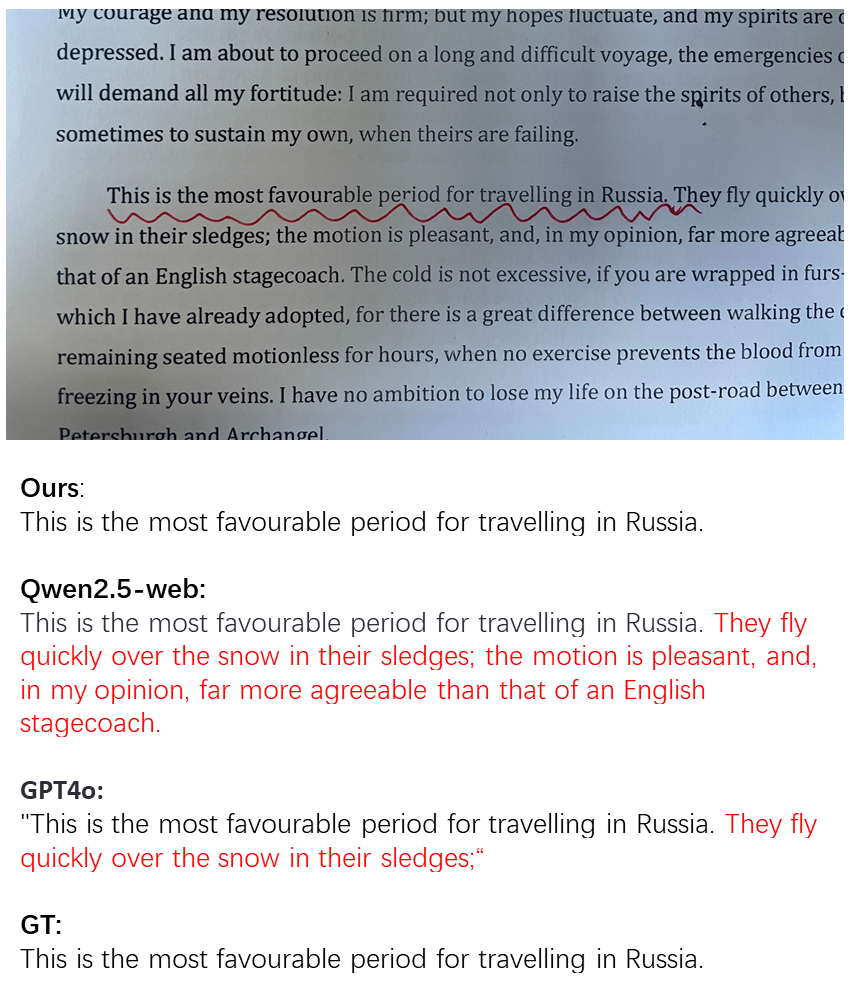}}
\caption{A example of in-wild data results.}
\label{fig:wild2}
\end{figure}

\begin{figure}[htbp]
\centerline{\includegraphics[width=0.5\textwidth]{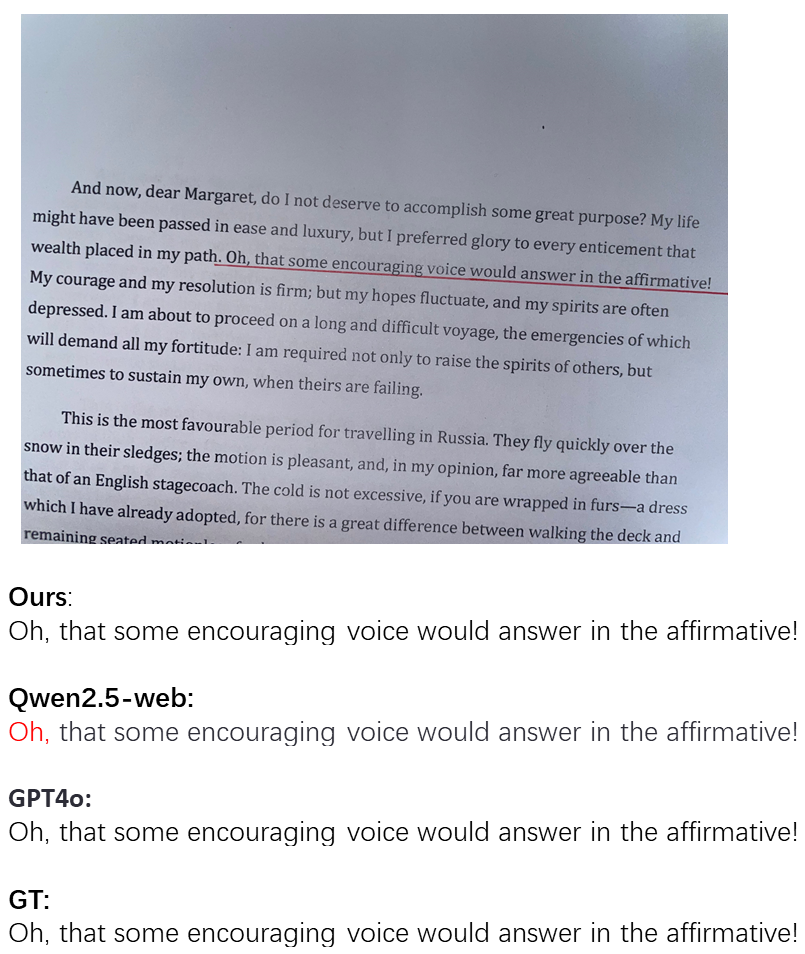}}
\caption{A example of in-wild data results.}
\label{fig:wild3}
\end{figure}

\begin{figure}[t]
\centerline{\includegraphics[width=0.5\textwidth]{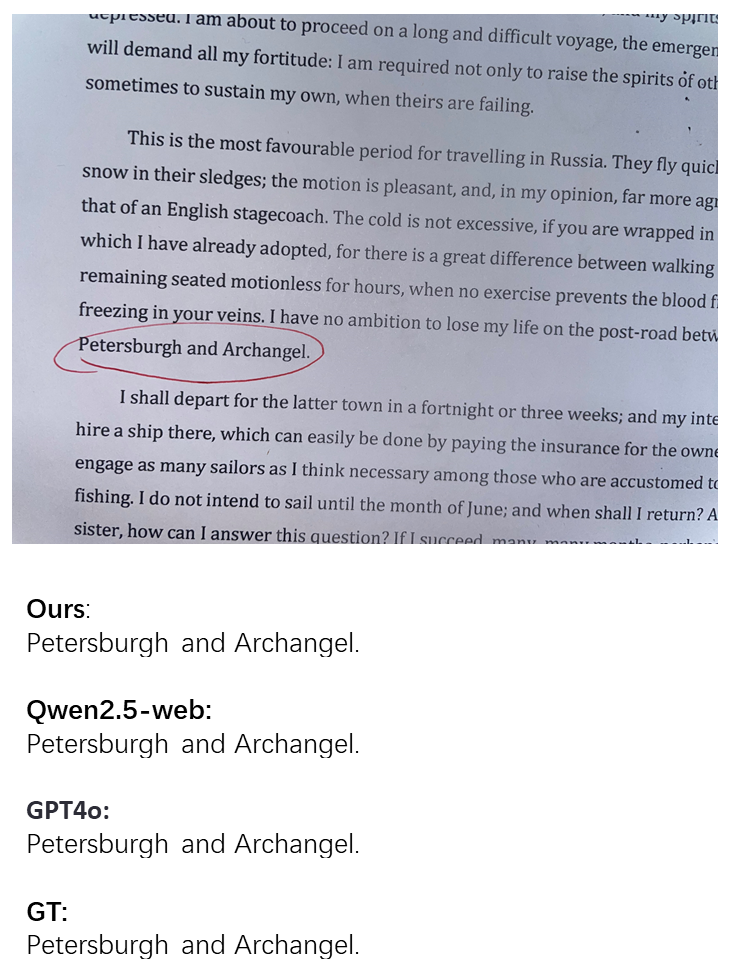}}
\caption{A example of in-wild data results.}
\label{fig:wild4}
\end{figure}

We also tested the performance of our model in real-world scenarios and compared it with the two best-performing models from our experimental comparison: GPT-4o and Qwen 2.5. Both of these models are large commercial models deployed on web platforms.Additional real-world data can be found in Figures~\ref{fig:wild} ~\ref{fig:wild2} ~\ref{fig:wild3} ~\ref{fig:wild4}. 

The results, as shown in the figure, indicate that for the PM annotation method, the model's judgments are somewhat inaccurate; however, the output remains highly relevant, which is consistent with our experimental findings. In the presence of noise, both commercial models also struggle to handle the situation effectively, leading to excessive and irrelevant outputs. In contrast, the Rect annotation method, and similar methods, achieved the highest accuracy, as their annotation approach is the most precise.

\end{document}